\documentclass{article}

\usepackage{arxiv}

\usepackage[utf8]{inputenc} 
\usepackage[T1]{fontenc}    
\usepackage{hyperref}       
\usepackage{url}            
\usepackage{booktabs}       
\usepackage{amsfonts}       
\usepackage{nicefrac}       
\usepackage{microtype}      
\usepackage{lipsum}
\usepackage{rotating}
\usepackage{tabularx}
\usepackage{tabu}
\usepackage{acronym}
\usepackage{bigstrut}
\usepackage{amsfonts}
\usepackage{amsmath}
\usepackage{algorithmic}
\usepackage{array}
\usepackage{color}
\usepackage{svg}
\usepackage{dblfloatfix}
\usepackage{graphicx}
\usepackage{multirow}
\usepackage{hyperref}
\usepackage{bm}
\usepackage[english]{babel}
\usepackage{cite}
\usepackage{url}
\usepackage{booktabs}
\newcommand{\tabitem}{~~\llap{\textbullet}~~}

\def\boolsupplementary{0}

\begin{document}
\title{ECG-DelNet: Delineation of Ambulatory Electrocardiograms with Mixed Quality Labeling Using Neural Networks}
\author{Guillermo~Jimenez-Perez$^*$,~Alejandro~Alcaine,~and~Oscar~Camara%
\thanks{G. Jimenez-Perez$^*$ and O. Camara are with the PhySense research group, BCN-MedTech, Department of Information and Communication Technologies, Universitat Pompeu Fabra, 08018 Barcelona, Spain. A. Alcaine is with the Biomedical Research Networking Center in Bioengineering, Biomaterials and Nanomedicine (CIBER-BBN), 28029 Madrid, Spain and with Biomedical Signal Interpretation and Computational Simulation (BSICoS) group, Arag{\'o}n Institute of Engineering Research, IIS Arag{\'o}n, University of
Zaragoza, 50018 Zaragoza, Spain. (Correspondence e-mail: {\color{blue} \href{mailto:guillermo.jimenez@upf.edu}{guillermo.jimenez@upf.edu}}).}%
\thanks{}}

\maketitle

\begin{abstract}
    Electrocardiogram (ECG) detection and delineation are key steps for numerous tasks in clinical practice, as ECG is the most performed non-invasive test for assessing cardiac condition. State-of-the-art algorithms employ digital signal processing (DSP), which require laborious rule adaptation to new morphologies. In contrast, deep learning (DL) algorithms, especially for classification, are gaining weight in academic and industrial settings. However, the lack of model explainability and small databases hinder their applicability. We demonstrate DL can be successfully applied to low interpretative tasks by embedding ECG detection and delineation onto a segmentation framework. For this purpose, we adapted and validated the most used neural network architecture for image segmentation, the U-Net, to one-dimensional data. The model was trained using PhysioNet's QT database, comprised of 105 ambulatory ECG recordings, for single- and multi-lead scenarios. To alleviate data scarcity, data regularization techniques such as pre-training with low-quality data labels, performing ECG-based data augmentation and applying strong model regularizers to the architecture were attempted. Other variations in the model's capacity (U-Net’s depth and width), alongside the application of state-of-the-art additions, were evaluated. These variations were exhaustively validated in a 5-fold cross-validation manner. The best performing configuration reached precisions of 90.12\%, 99.14\% and 98.25\% and recalls of 98.73\%, 99.94\% and 99.88\% for the P, QRS and T waves, respectively, on par with DSP-based approaches. Despite being a data-hungry technique trained on a small dataset, DL-based approaches demonstrate to be a viable alternative to traditional DSP-based ECG processing techniques.
\end{abstract}


\section{Introduction}\label{sec:introduction}

Surface electrocardiogram (ECG) is the main cardiac diagnostic and monitoring tool in clinical practice due to its widespread accessibility, ease of use and the rich representation of relevant structural and functional information in its waveform. Usually, physicians perform visual inspection of the ECG, often manually delineating the QRS complex, in order to diagnose a patient, interpreting and evaluating potential pathological deviations in the waveform. However, these markers might go unnoticed to non-specialists or even to trained cardiologists, especially when analysing multiple leads simultaneously for several heart cycles or in stress-related situations such as in the intensive care unit.

Computational methods can help unburden physicians of these problems by providing objective measurements over clinical data \cite{Minchole2019, Minchole2019a} or by aiding in the discovery of potential biomarkers, finding hidden patterns of statistical relevance \cite{Lyon2018, Faust2018}. For these purposes, ECG detection and delineation (hereinafter delineation) is often a prerequisite step, aiding in data structuring prior to any downstream task \cite{Lyon2018}. ECG delineation consists in computing fiducials for each of the different ECG waves (P, QRS and T waves), delimiting the starting (onset) and ending (offset) points of each wave. If multiple leads are available, delineation can be performed directly on all available leads (multi-lead) or on the individual leads for their posterior aggregation (single-lead). However, although more conceptually straightforward, the input heterogeneity of multi-lead strategies increases: clinical practice employs configurations ranging from a single lead (wearables) to ~250 leads (ECG imaging) with various placements, needing higher amounts of annotated data for algorithm training and validation.

Several computational methods exist in the literature for processing ECG data. Although digital signal processing (DSP) algorithms have historically been used for this purpose, the machine learning (ML) community has recently developed tools to work with ECGs, using both ``classical'' approaches and deep learning (DL). ML-based methods for ECG analysis are, however, relatively scarce when compared to other fields in biomedical engineering and have mainly focused on classification \cite{Hou2019}, especially in the case of DL-based approaches. As suggested by Pinto et al. in their recent review of ECG-based biometrics \cite{Pinto2018}, data-driven techniques based in DL suffer from the lack of large and manually annotated ECG databases, which usually include less than a hundred patients. Other works have employed 2-dimensional neural networks over the ECG spectrogram for classification purposes, bypassing the natural one-dimensional representation \cite{Huang2019}.

To the best of our knowledge, and partially due to the reasons exposed above, DSP algorithms using the wavelet transform and rule-based adaptive thresholds remain state of the art for ECG delineation \cite{Li1995, Martinez2004, Banerjee2012}, reaching high precision and recall values of over 95\% for all delineated waves. However, these methods require laborious rule adaptation when extended to scenarios where they underperform; these algorithms have been developed using the whole dataset for setting the rules, compromising their generalization on unseen waveforms. 

Classical ML algorithms, namely Gaussian mixture models \cite{Dubois2007} or hidden Markov models (HMM) \cite{Graja2005}, have been applied to delineation. However, these methods might scale poorly when trained on large-scale databases that represent a wider variety of pathologies, as well as presenting sensitivity reductions with respect to DSP- and DL-based algorithms. DL techniques have also been used for delineation in the shape of convolutional neural networks (CNN) \cite{Camps2019, Sodmann2018}, long short-term memory (LSTM) networks \cite{Hedayat2018} and fully-convolutional networks (FCN) \cite{Tison2019, Jimenez-Perez2019}. However, some of these works perform delineation solely on the QRS wave \cite{Camps2019}, whereas others validate their performance uniquely on sinus rhythm \cite{Tison2019} or show reduced performance compared to DSP-based approaches \cite{Hedayat2018, Sodmann2018, Jimenez-Perez2019}. Furthermore, CNN-based models are non-translation invariant \cite{Camps2019, Sodmann2018}, as they perform predictions using fully connected layers, property that we consider key for robust delineation \cite{Garcia-Garcia2018}. In addition, no DL-based approach published up to date propose any data management technique (data augmentation, semi-supervised training), either ECG-tailored or otherwise.

In this work we present the adaptation of the classical U-Net architecture \cite{Ronneberger2015}, the most successful type of FCN currently used in image segmentation, for single-lead and multi-lead ECG fiducial inference. The developed methodology was tested on the PhysioNet's QT database \cite{Laguna1997}, which holds approximately 3,000 two-lead beats annotated by expert cardiologists having both leads in sight. For its usage in this scenario, the U-Net had to be adapted to work with one-dimensional (signal) data, framing delineation as a segmentation task. The dataset was expressed as binary masks for the utilization the Dice score, the usual loss employed in image segmentation tasks.

Several regularization strategies were also applied to successfully train the delineator for overcoming the difficulties posed by the relatively small dataset, ECG information redundancy and large inter- and intra-patient variability. The regularization techniques consisted in developing ECG-tailored data augmentation (DA) transformations such as amplifier saturation or powerline noise for improving the network's generalizability, in pre-training the model with low-quality labeled data as ground truth and in adding in-built regularizers such as Spatial Dropout (SDo) alongside Batch Normalization (BN) in the architecture. Other architectural modifications such as the application of Atrous Spatial Pyramid Pooling (ASPP) \cite{Chen2018a}, hyper-dense connectivity (HdC) \cite{Dolz2018} and multi-scale upsampling (MsU) \cite{Chen2018a} were also explored in this work.

The rest of the paper is organized as follows. Section \ref{sec:materials} describes the employed database. Section \ref{sec:methods} details the methodology followed in this work. Section \ref{sec:results} addresses the results obtained by this work. Section \ref{sec:discussion} discusses about the obtained results and their implications on the feasibility of applying this pipeline in the clinical practice. Finally, Section \ref{sec:conclusions} summarizes this work's conclusions. A preliminary version of this work has been reported in \cite{Jimenez-Perez2019}.

\section{Materials}\label{sec:materials}

\begin{figure}[!t]
    \centering
    \setcounter{figure}{0}
    \includegraphics[width=0.9\linewidth]{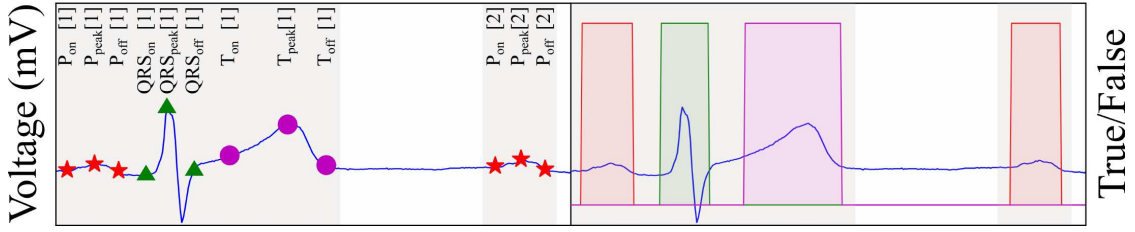}
    \caption{\label{fig:fiducials} Database's high-quality ground truth as fiducials (left) and as three overlapped binary masks (right). Red mask/stars: P wave. Green mask/triangles: QRS wave. Magenta mask/circles: T wave. The figure depicts the employed notation.}
\end{figure}

The QT database was employed for model training and evaluation \cite{Laguna1997}. The database is comprised of $105$ ambulatory, two-lead recordings of 15 minutes of length at sampling frequency ($f_s$) of 250 Hz. A variety of pathologies are represented in the database, including ventricular and supraventricular arrhythmia, ischemic and non-ischemic ST episodes, slow ST level drift, transient ST depression and sudden cardiac death. Two label sets exist per recording: a high-quality annotation performed by a cardiologist (``high-quality'', hereinafter) consisting of approximately $30$ fully delineated beats per recording, and an automatic delineation (``low-quality'') performed using the ECGpuwave algorithm on every beat of each recording \cite{Goldberger2000}. The low-quality ground truth is produced in a single-lead manner, whereas the high-quality dataset is annotated in a multi-lead fashion, producing a single set of markers for both leads. When using the low-quality ground truth for multi-lead predictions, the lead with a better Dice coefficient with respect to the high-quality ground truth was used. Each annotation set holds (at most, if the P wave is present) nine fiducials per beat: the P, QRS and T wave detection markers and their respective onsets and offsets. 

Some recordings in the high-quality dataset had to be partially re-annotated, as they contained extrasystolic beats that were neither detected nor delineated. Specifically, 112 beats in recordings \textit{sel102}, \textit{sel213}, \textit{sel221}, \textit{sel308}, \textit{sel44} and \textit{sel820} were added. The correction was necessary as our algorithm requires fully delineated windows as input while training. If non-delineated beats exist within a window, the training procedure might find inconsistent parameters. Isolated delineations in the high-quality dataset were also excluded, as they were unusable for training the algorithm. Records \textit{sel232}, \textit{sel233} and \textit{sel36} were discarded given that the annotations were incomplete. A single recording, \textit{sel35}, was discarded due to being the only recording in atrial flutter, making it impossible to abstract this morphology with a single example. The final database used in our work contained a total of $3,246$ beats with both high- and low-quality labels and $131,924$ beats for which only low-quality labeling was available.

Throughout the rest of the paper the following notation will be used: recordings will be denoted as $\mathcal{R} = [r_1, r_2, ..., r_{105}]$; each recording $r$ has two leads $L^{(r)} = [\ell^{(r)}_1, \ell^{(r)}_2]$; waves $W$ (either P, QRS or T) present in the $\ell$-th lead of a recording $r$ are the set of their fiducials $W^{(r,\ell)} = \{\textbf{w}^{(r,\ell)}_{\textrm{on}},\textbf{w}^{(r,\ell)}_{\textrm{peak}}, \textbf{w}^{(r,\ell)}_{\textrm{off}}\}$, and where $\textbf{w}^{(r,\ell)}_{\textrm{i}}$ encodes any the possible set of fiducials; and fiducials $\textbf{w}^{(r,\ell)}_{\textrm{i}} =  \left[w^{(r,\ell)}_{\textrm{i}}[1],...,w^{(r,\ell)}_{\textrm{i}}[M]\right]$, encode the samples of occurrence of all fiducials $M$ present in a lead of a recording, where $w^{(r,\ell)}_{\textrm{i}}[i] \in [0,N]$ and $N$ is the total number of samples per recording. Thus, the 25th T wave onset on the 1st lead of the 4th recording is noted as ``$T^{(4,1)}_{\textrm{on}}[25]$''. An example can be seen in Figure \ref{fig:fiducials}.

\section{Methods}\label{sec:methods}

\begin{figure}[!t]
    \centering
    \includegraphics[width=\linewidth]{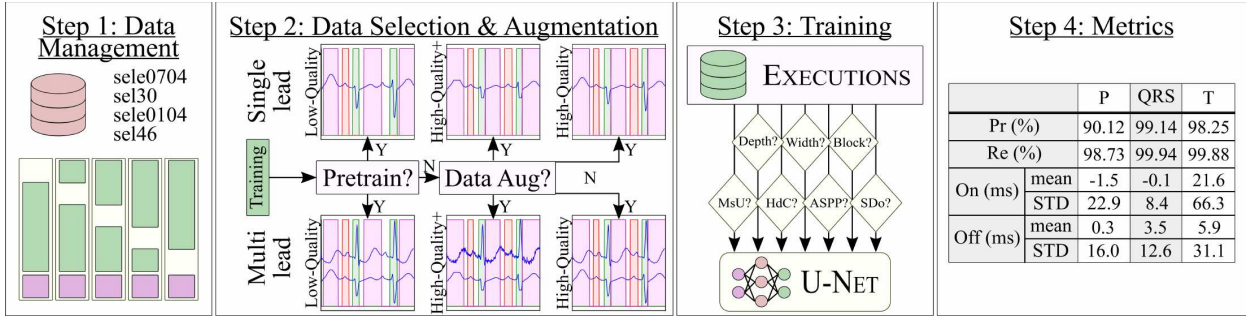}
    \caption{\label{fig:pipeline}
    Developed DL-based methodology for ECG delineation. Step 1: the randomly shuffled data is first split into train and test sets (green and purple blocks, respectively) in a 5-fold cross-validation manner. Step 2: the cardiologist-annotated dataset (high-quality) is then prepared for training, either as single-lead or multi-lead, with possible pre-training with low-quality labels or ECG-tailored data augmentation (high-quality+). Step 3: for each fold, the network is trained and the test set is subsequently predicted. The base network, the U-Net, is instantiated and modified with state-of-the-art architectural modifications drawn from the literature, such as multi-scale upsampling (MsU), hyper-dense connectivity (HdC), atrous spatial pyramid pooling (ASPP) and spatial dropout (SDo). Step 4: the performance of the model is evaluated by comparing the high-quality and predicted labels for all folds. The employed metrics for evaluation are the prediction (Pr), recall (Re) for detection and the mean and standard deviation (STD) of the onset (On) and offset (Off) markers.}
\end{figure}

The developed DL-based methodology for ECG delineation is depicted in Figure \ref{fig:pipeline}. The first step describes the data splitting and management (Section \ref{sec:datamanagement}). The second step outlines the dataset selection and the developed data augmentation (Section \ref{sec:dataselandaug}). The third step summarizes the base architecture and its additions (Sections \ref{sec:basearch} and \ref{sec:archvariations}). The fourth step details the evaluation methodology (Section \ref{sec:evaluation}). Finally, the tested configurations are listed in Section \ref{sec:experiments}. We have made our code publicly available in \url{https://github.com/guillermo-jimenez/ECGDelNet}.

\subsection{Data Management}\label{sec:datamanagement}

In ML procedures, data instances are usually divided into train, test and validation sets in a non-overlapping manner. However, ECG signals are recordings of the electrical activity of several beats as captured by different leads that are usually windowed for reducing input complexity. This windowing alongside the existence of simultaneous views of the data provided by the leads increases the risk of performing an incorrect splitting of the dataset, assigning similar representations of the same entity to different sets. If the splitting ``instance'' is defined at the window/beat level, virtually identical representations of the data could be included in both the training and testing sets. If it is defined at the view/lead level, highly correlated information can be assigned to different sets. Models trained with this flawed splitting incur the risk of memorizing the data instead of inferring abstract patterns over it, especially in the case of high capacity models such as DL. According to Faust \textit{et al}. \cite{Faust2018}, although undesirable, this practice is widespread in ECG-based machine learning procedures. Following their recommendations, we performed subject-wise splitting, comprising all windows of both leads and DA (if applicable), in a 5-fold cross-validation manner. 

\subsection{Data Selection and Augmentation}\label{sec:dataselandaug}

Given the relatively small amount of manually tagged data, with $\sim3000$ beats and low intra-recording per-beat variability, some decisions were made with respect to data handling. On the first hand, two inference strategies were attempted for delineation: single-lead and multi-lead. When using single-lead annotations, the algorithm would be fed one lead at a time, producing a mask for every lead separately. When using multi-lead annotations, the network receives both leads in the recording as input, producing a single mask as an output. Secondly, three different training strategies were attempted: training with high-quality data alone, pre-training the algorithm with low-quality data obtained by a DSP-based algorithm \cite{Goldberger2000} and applying a custom DA scheme. 

Lastly, for using the U-Net architecture, all fiducials $W_i^{(r,\ell)}$ for a recording $r$ and lead $\ell$ were transformed into Boolean masks:

\vspace{-0.5em}
\begin{equation}\label{eqn:mask}
    B^{(r,\ell)}[n] = \left\lbrace \begin{array}{ll}
        1 & \textrm{if} \; n \in \left[w^{(r,\ell)}_{\textrm{on}}[m], w^{(r,\ell)}_{\textrm{off}}[m]\right]_{0..M} \\
        0 & \textrm{otherwise} \\
    \end{array} \right.,
\end{equation}
where $B$ is the produced binary mask, $m \in [0,M]$ are the ground truth fiducials, and $n$ is the signal's sample number. Figure \ref{fig:fiducials} depicts both the original fiducials and the binary masks.

Data augmentation improves a network's generalization by adding realistic noise sources to the input data, learning noise-insensitive representations \cite{Perez2017} and acting as a regularizer. In this work, six different noise sources, computed to have a specific signal-to-noise ratio (SNR) with respect to one input signal, were developed and specifically tailored to ECGs, comprising additive white Gaussian noise (AWGN), random periodic spikes (RS), amplifier saturation (AS), pacemaker spikes (PS), powerline noise (PN) and baseline wander (BW):

\begin{equation*}
    AWGN[n] = \mathcal{N}\left(0,\sqrt{\tilde{P}_n}\right)
\end{equation*}
\begin{equation*}
    RS[n] = \sqrt{\frac{\tilde{P}_n}{f}} \sum\limits_{k = -\infty}^{k = \infty} (\delta * Sp)\left[n - k \frac{1}{f}\right]
\end{equation*}
\begin{equation}
    AS[n] = \left\lbrace \begin{array}{ll} - x[n] + S_v & if \, x[n] \geq S_v \\ - x[n] - S_v & if \, x[n] \leq - S_v \\ 0 & otherwise \end{array} \right.
\end{equation}
\begin{equation*}
    PS[n] = \left\lbrace \begin{array}{ll} \sqrt{\frac{\tilde{P}_n}{f}} & if\,n \in QRSon \\ 0 & otherwise \end{array} \right.
\end{equation*}
\begin{equation*}
    PN/BW[n] = \sqrt{2 \tilde{P}_n} \cos{\left(\frac{2\pi f}{f_s} n\right)},
\end{equation*}
where $\mathcal{N}$ is the normal distribution, $\tilde{P}_n = P_s/10^{\textrm{SNR}/10}$ is the noise power, $P_s$ is the input signal power, $f_s$ is the sampling frequency, $S_v = p  \max{\left| \textbf{x} \right|}$ is the saturation value, $(a * b)[n]$ indicates the convolution operation, $Sp = [0,0.15,1.5,-0.25,0.15]^T + \mathcal{U}(-0.25, 0.25)$ is a custom filter with uniform noise that models pacemaker spikes and $\delta$ is the impulse function.

\begin{figure}[!t]
    \centering
    \includegraphics[width=1\linewidth]{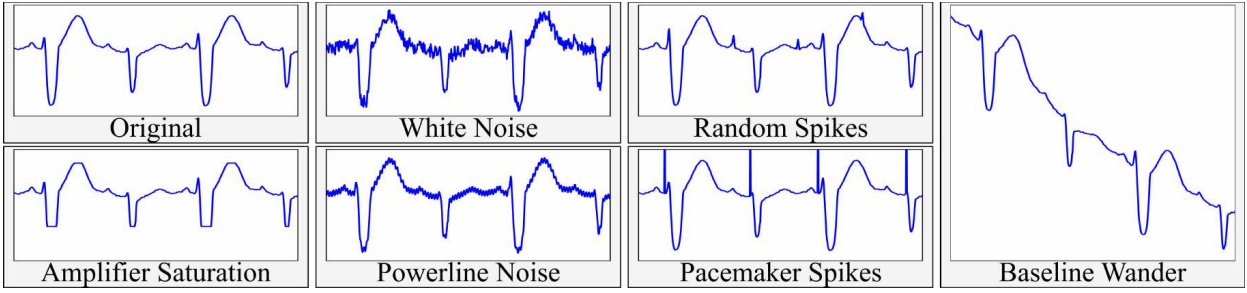}
    \caption{\label{fig:dataaugmentation} 
    Data augmentation strategy example for an ECG recording. Re-execution results in slight signal-to-noise ratio (SNR) and frequency ($f$) variations, altering the final shape of the computed noise.}
\end{figure}

The first five noise sources were engineered to represent the observed variations in the dataset. Pacemaker spikes were designed to avoid misidentifying spike-like noise near QRS complexes and for completeness. The values of the noise-generating parameters were altered from window to window for flexibility, given $p^{(i)} = p + \mathcal{U}(\pm SNR/10)$, where $\mathcal{U}$ is the uniform distribution. Figure \ref{fig:dataaugmentation} depicts an example of the employed sources.

\subsection{U-Net Architecture}\label{sec:basearch}

\begin{figure}[!b]
    \centering
    \includegraphics[width=1\linewidth]{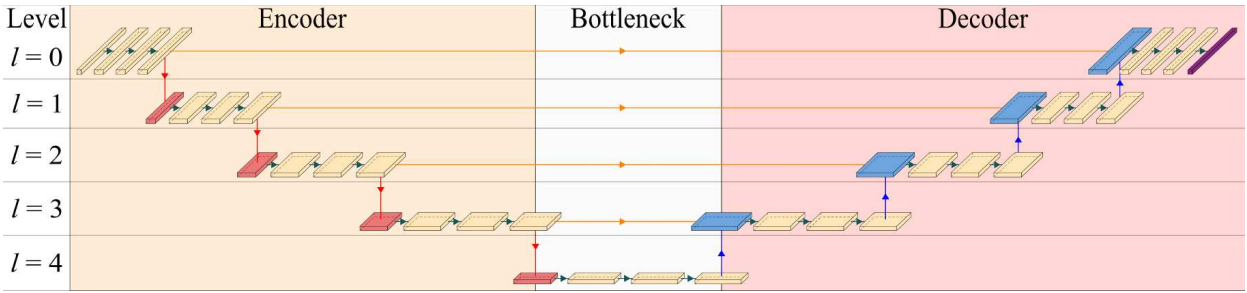}
    \caption{\label{fig:architecture}
    Base U-Net instantiated with 5 levels and 3 convolutional blocks per level. All blocks represent tensors outputted by specific operations: convolutions (yellow), pooling (red), upsampling followed by skip connections (blue) and convolution with sigmoid activation function (purple). All convolutional blocks extract $2^\textit{l}  N$ channels, whereas pooling and upsampling have a size of $2$.}
\end{figure}

This work employs the U-Net \cite{Ronneberger2015} as its base architecture, consisting in an encoder, a bottleneck and a decoder with skip connections between the encoder and the decoder, as can be appreciated in Figure \ref{fig:architecture}. The encoder extracts increasingly abstract representations of the input data through several levels of stacked convolutional operations and downsampling blocks. The decoder recovers information from the bottleneck, the nexus between the encoder and the decoder, through convolutional and upsampling blocks. Skip connections allow for finer segmentation at the object boundaries by direct information transmission from the encoder. In U-Net networks and related architectures the number of convolutional filters is usually doubled after each downsampling block and halved after each upsampling block from its initial $N$ channels.

For a clearer exposition, we have grouped operations in the U-Net architecture into ``blocks'', forming ``levels''. We define a ``block'' as an ordered composition of operations on a tensor $\textbf{x}$. In the same line, we define a ``level'' as a set of operations whose results have compatible tensor size. The considered blocks are convolutional, downsampling, upsampling, and skip connection blocks.

Convolutional blocks are the main operation in the U-Net, consisting in a combination of convolutional operations $C(\cdot)$ (or separable convolutions $S(\cdot)$), ReLU non-linearities $N(\cdot)$, regularizers $R(\cdot)$ and, optionally, point-wise additions $A(\cdot, \cdot)$. Following state-of-the-art practices working with U-Net architectures, we explored the following convolutional blocks:

\begin{table}[ht]
    \centering
    \normalsize
    \begin{tabular}{lll}
        \tabitem Vanilla  & $\textbf{y} = \;\;\;\;C(R(N(C(R(N(\textbf{x}))))))$ & \cite{Ronneberger2015} \\
        \tabitem Residual & $\textbf{y} = A(C(R(N(C(R(N(\textbf{x})))))), \textbf{x})$ & \cite{He2016b} \\
        \tabitem XCeption & $\textbf{y} = A(S(R(N(S(R(N(\textbf{x})))))), \textbf{x})$ & \cite{Chollet2017} 
    \end{tabular}
\end{table}

Upsampling blocks perform bilinear interpolation on a tensor, whereas downsampling blocks perform an averaging of neighboring samples. These operations do not alter the number of channels but the tensor size. Lastly, skip connections perform a concatenation of the result of the upsampling operations in the decoder and the last tensor of the encoder at each level. The ordering of the blocks and the number of convolutional operations per block were defined to agree with the literature \cite{Chollet2017, He2016b}. 

Several modifications to the original U-Net were required for its adaptation to ECG analysis. Firstly, we replaced the original 2D convolutions by 1D operations, more suited for signal processing. Secondly, our network applies zero-padding, maintaining the tensor's shape after each convolutional operation and avoiding information loss at the boundaries. Thirdly, we included a stem (one extra convolutional module right after the input) mimicking classification architectures \cite{LeCun1998, Szegedy2014, Chollet2017}.

\subsection{Variations over Base Architecture}\label{sec:archvariations}

The U-Net is the base for different optional additions taken from state-of-the-art architectures in the literature. Given an initial model testing phase, the need for strong regularization techniques other than batch normalization were apparent. We opted to apply SDo \cite{Tompson2015}, which consists in randomly dropping entire channels of different convolutional operations in the training phase, as opposed to standard dropout, where specific neurons are dropped in an unstructured manner. 

\begin{figure}[!t]
    \centering
    \includegraphics[width=1\linewidth]{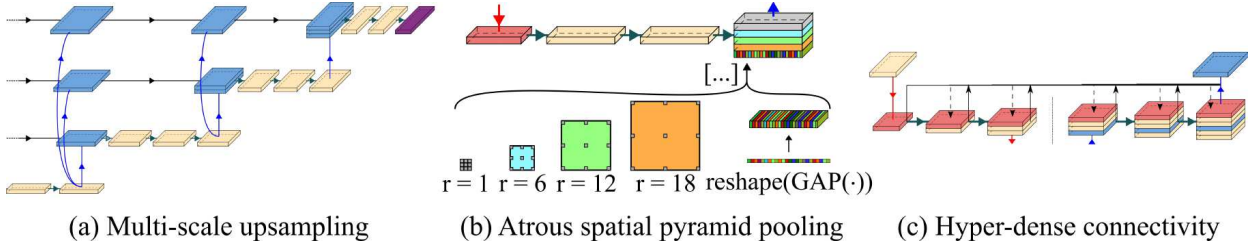}
    \caption{\label{fig:variations}
    Architectural variations. (a) Multi-scale upsampling. (b) Atrous spatial pyramid pooling module. (c) Hyper-dense connectivity. Blocks represent output tensors: convolutions (yellow), pooling (red), upsampling $+$ skip connections (blue), convolution with sigmoid activation (purple). Arrows represent operations: convolution (green), concatenation (black), skip connections (orange), downsampling (red) and upsampling (blue). GAP stands for global average pooling. $r$ represents the dilation rate of the atrous convolution.}
\end{figure}

Other variations included multi-scale upsampling (MsU) \cite{Chen2018a}, atrous spatial pyramid pooling (ASPP) \cite{Chen2018a} and hyper-dense connectivity (HdC) \cite{Dolz2018}, taken from high-ranking image segmentation architectures. MsU attempts at directly propagating abstract information in deeper layers by upsampling the last tensor in each level of the decoder and concatenating it alongside the skip connection to the immediately shallower level, providing general context to the output level (Figure \ref{fig:variations}-a). ASPP attempts at retrieving global contextual information at the deepmost level in the architecture by concatenating tensors resulting from parallely applying \textit{atrous} (or dilated) convolutions with different dilation rates $r$ and a image pooling operation to the same input (Figure \ref{fig:variations}-b). HdC \cite{Dolz2018} generalizes residual \cite{He2016b} and dense \cite{Huang2017} connections to an arbitrary number of tensors with compatible shapes to minimize filter redundancy by concatenating all output tensors from previous blocks with compatible input shape and using them as input to the next operation, aiming at better discrimination (Figure \ref{fig:variations}-c).

\subsection{Evaluation}\label{sec:evaluation}

The evaluation of the proposed methodology is inspired by the metrics used in state-of-the-art DSP-based algorithms \cite{Martinez2004} for comparison purposes. A binary mask of the correspondence between the true fiducials, $W^{(r,\ell)}$, and the predicted fiducials, $\hat{W}^{(r,\ell)}$, for a recording $r$ and a lead $\ell$ can be computed as:

\begin{equation}\label{eqn:boolmask}\arraycolsep=1.4pt
    H_{jk}^{(r,\ell)} = 
    \left\lbrace
    \begin{array}{lll} 
        1 & \textrm{if} & (\hat{w}_{\textrm{i}}^{(r,\ell)}[k] \in [w_{\textrm{on}}^{(r,\ell)}[j], w_{\textrm{off}}^{(r,\ell)}[j]]) \\
        & \textrm{or} & (w_{\textrm{i}}^{(r,\ell)}[k] \in [\hat{w}_{\textrm{on}}^{(r,\ell)}[j], \hat{w}_{\textrm{off}}^{(r,\ell)}[j]]) \\
        0 & \multicolumn{2}{l}{\textrm{otherwise}}
    \end{array}
    \right.,
\end{equation}
where $j \in [0,M]$ and $k \in [0,\hat{M}]$ are the total true and predicted fiducials, respectively. 

The individual lead information is then combined into a single mask through an logical ``OR'' operator $\bar{H} = OR(H^{(0)}, ..., H^{(L)})$ so that the true positives (TP) for a given recording $r$ are $TP_r = \sum \bar{H}_{jk}$. False-positives (FP), on their behalf, are elements of $\tilde{\textbf{w}}_i^{(r,\ell)}$ that did not correspond to any true fiducial ($FP_r = \hat{M} - card(\{(j,k) \mid \bar{H}_{jk} = 1 \})$). Finally, a false-negative (FN) is considered when the ground truth displays a beat that is not captured by a TP (corresponding to $FN_r = M - card(\{(j,k) \mid \bar{H}_{jk} = 1 \})$). The precision ($Pr$) and recall ($Re$) for each $r$ and $\ell$ were computed, reporting in this work the overall performance for all recordings and leads for conciseness. Additionally, the F1 score is also computed for comparing different architectural variations, as a single figure of merit.

The delineation metrics were computed for the TPs (cases where $\bar{H}_{jk} = 1$), as no onset/offset correspondences between the GT and the prediction exist otherwise. The relative error of the segmentation was computed through the mean and standard deviation (SD) of the difference of the actual and predicted onsets or offsets of the correspondences found in Eq. \ref{eqn:boolmask}:

\begin{equation}
    \begin{aligned}
    \min_{j, k, \ell} \quad & w_i^{(r,\ell)}[j] - \hat{w}_i^{(r,\ell)}[k] \\
    \textrm{s.t.} \quad & \bar{H}^{(\ell)}_{jk} = 1.
    \end{aligned}
\end{equation}

\subsection{Experiments}\label{sec:experiments}

A series of variations on the data level and in the network's topology were attempted in this work to test the robustness of the model. The data-level variations (described in Section \ref{sec:dataselandaug}) were aimed at alleviating data scarcity of the QT database and consisted in applying the in-built SDo regularization technique alongside batch normalization (with 25\% dropout rate, commonly found in the literature), DA strategies and pre-training with low-quality labelling of the dataset for either single- or multi-lead inference strategies. Given computational constraints, no DA was applied when low-quality pre-training was performed.

For all of the above variations, a raid of possible architectural changes were independently tested. Firstly, the benefit of applying MsU, ASPP or HdC was assessed to account for changes in model capacity and expressiveness. Secondly, the type of convolutional block employed (``vanilla'', residual, XCeption), the network's depth ($L \in [4, 5, 6, 7]$), the number of convolutional blocks per level ($CB \in [ 2, 3, 4, 5, 6 ]$) were attempted. In total, 201 executions were performed testing various configurations, with training times ranging from 6 hours to several days. The executions were performed in a high performance computing environment where each configuration was assigned to a single NVIDIA 1080Ti or NVIDIA Titan Xp GPU. To ensure result comparability, the same random seed was employed in all executions.

Some aspects were kept constant throughout the executions, such as the nonlinearity (ReLU for all blocks and sigmoid for the last block), the kernel size (3), the loss function (Jaccard), the optimizer (Adam \cite{Kingma2015}) and the random seed (1234).

\section{Results}\label{sec:results}

\subsection{Model selection}

\begin{table*}[t!]
\centering
{
\scriptsize
\begin{tabular}{cccccccccc}
\cline{2-10}
\multicolumn{1}{c|}{}        
    & \multicolumn{3}{c||}{Spatial Dropout}
    & \multicolumn{3}{c||}{Pre-train with low-quality Labels}
    & \multicolumn{3}{c|}{Data Augmentation}
    \\ \hline
\multicolumn{1}{|c|}{Metric} 
    & \multicolumn{1}{c|}{$\Delta$F1} 
    & \multicolumn{1}{c|}{\begin{tabular}[c]{@{}c@{}}$\Delta$Onset \\ M$\pm$SD (ms)\end{tabular}} 
    & \multicolumn{1}{c||}{\begin{tabular}[c]{@{}c@{}}$\Delta$Offset \\ M$\pm$SD (ms)\end{tabular}} 
    & \multicolumn{1}{c|}{$\Delta$F1}   
    & \multicolumn{1}{c|}{\begin{tabular}[c]{@{}c@{}}$\Delta$Onset \\ M$\pm$SD (ms)\end{tabular}} 
    & \multicolumn{1}{c||}{\begin{tabular}[c]{@{}c@{}}$\Delta$Offset \\ M$\pm$SD (ms)\end{tabular}} 
    & \multicolumn{1}{c|}{$\Delta$F1}   
    & \multicolumn{1}{c|}{\begin{tabular}[c]{@{}c@{}}$\Delta$Onset \\ M$\pm$SD (ms)\end{tabular}} 
    & \multicolumn{1}{c|}{\begin{tabular}[c]{@{}c@{}}$\Delta$Offset \\ M$\pm$SD (ms)\end{tabular}} 
    \\ \hline
\multicolumn{1}{|c|}{P}      
    & \multicolumn{1}{c|}{+ \textbf{1.98} \%}  
    & \multicolumn{1}{c|}{$-$2.79 $-$ 0.83}
    & \multicolumn{1}{c||}{$-$2.07 $-$ 1.18}
    & \multicolumn{1}{c|}{+ 1.85 \%}    
    & \multicolumn{1}{c|}{$-$\textbf{0.18} $-$ \textbf{4.31}}
    & \multicolumn{1}{c||}{$-$0.64 $-$ 2.51}
    & \multicolumn{1}{c|}{+ 1.22 \%}    
    & \multicolumn{1}{c|}{+0.86 $-$ 2.10}
    & \multicolumn{1}{c|}{$-$\textbf{0.42} $-$ \textbf{3.23}}
    \\ \hline
\multicolumn{1}{|c|}{QRS}    
    & \multicolumn{1}{c|}{+ \textbf{3.27} \%}
    & \multicolumn{1}{c|}{$-$\textbf{0.73} $-$ \textbf{1.43}}
    & \multicolumn{1}{c||}{+\textbf{1.88} $-$ \textbf{2.90}}
    & \multicolumn{1}{c|}{+ 1.07 \%}
    & \multicolumn{1}{c|}{+0.08 $-$ 0.74}
    & \multicolumn{1}{c||}{$-$0.23 $-$ 1.56}
    & \multicolumn{1}{c|}{+ 0.63 \%}
    & \multicolumn{1}{c|}{+0.26 $-$ 0.90}
    & \multicolumn{1}{c|}{+0.61 $-$ 1.01}
    \\ \hline
\multicolumn{1}{|c|}{T}
    & \multicolumn{1}{c|}{+ \textbf{7.21} \%}
    & \multicolumn{1}{c|}{+8.06 + 3.90}
    & \multicolumn{1}{c||}{$-$2.81 $-$ 1.32}
    & \multicolumn{1}{c|}{+ 1.65 \%}
    & \multicolumn{1}{c|}{+\textbf{1.53} $-$ \textbf{3.88}}
    & \multicolumn{1}{c||}{$-$\textbf{0.36} $-$ \textbf{4.87}}
    & \multicolumn{1}{c|}{+ 0.80 \%}
    & \multicolumn{1}{c|}{+3.70 $-$ 0.39}
    & \multicolumn{1}{c|}{$-$0.90 + 3.63}
    \\ \hline

&&&&&&&&&\\ \cline{2-10} 

\multicolumn{1}{c|}{}        
    & \multicolumn{3}{c||}{Multi-scale upsampling (\textendash SDo)}
    & \multicolumn{3}{c||}{Atrous spatial pyramid pooling (\textendash SDo)}
    & \multicolumn{3}{c|}{Hyper-dense connectivity (\textendash SDo)}
    \\ \hline
\multicolumn{1}{|c|}{Metric} & \multicolumn{1}{c|}{$\Delta$F1} 
    & \multicolumn{1}{c|}{\begin{tabular}[c]{@{}c@{}}$\Delta$Onset\\ M$\pm$SD (ms)\end{tabular}}
    & \multicolumn{1}{c||}{\begin{tabular}[c]{@{}c@{}}$\Delta$Offset\\ M$\pm$SD (ms)\end{tabular}}
    & \multicolumn{1}{c|}{$\Delta$F1}   
    & \multicolumn{1}{c|}{\begin{tabular}[c]{@{}c@{}}$\Delta$Onset\\ M$\pm$SD (ms)\end{tabular}}
    & \multicolumn{1}{c||}{\begin{tabular}[c]{@{}c@{}}$\Delta$Offset\\ M$\pm$SD (ms)\end{tabular}}
    & \multicolumn{1}{c|}{$\Delta$F1}   
    & \multicolumn{1}{c|}{\begin{tabular}[c]{@{}c@{}}$\Delta$Onset\\ M$\pm$SD (ms)\end{tabular}}
    & \multicolumn{1}{c|}{\begin{tabular}[c]{@{}c@{}}$\Delta$Offset\\ M$\pm$SD (ms)\end{tabular}}
    \\ \hline
\multicolumn{1}{|c|}{P}      
    & \multicolumn{1}{c|}{+ \textbf{1.52} \%}  
    & \multicolumn{1}{c|}{+4.18 + 0.62}
    & \multicolumn{1}{c||}{$-$0.71 $-$ 2.03}
    & \multicolumn{1}{c|}{$-$5.08 \%}     
    & \multicolumn{1}{c|}{$-$5.69 + 7.36}
    & \multicolumn{1}{c||}{+2.06 + 0.04}
    & \multicolumn{1}{c|}{+ 1.13 \%}    
    & \multicolumn{1}{c|}{+\textbf{1.09} $-$ \textbf{6.40}}
    & \multicolumn{1}{c|}{$-$\textbf{1.63} $-$ \textbf{5.83}}
    \\ \hline
\multicolumn{1}{|c|}{QRS}    
    & \multicolumn{1}{c|}{+ 1.93 \%}
    & \multicolumn{1}{c|}{+3.81 $-$ 1.07}
    & \multicolumn{1}{c||}{$-$\textbf{1.93} $-$ \textbf{3.51}}
    & \multicolumn{1}{c|}{+ 1.56 \%}
    & \multicolumn{1}{c|}{$-$\textbf{1.17} $-$ \textbf{2.03}}
    & \multicolumn{1}{c||}{+2.51 $-$ 3.61}
    & \multicolumn{1}{c|}{+ \textbf{2.21} \%}
    & \multicolumn{1}{c|}{+2.70 + 2.00}
    & \multicolumn{1}{c|}{$-$3.86 + 16.45}
    \\ \hline
\multicolumn{1}{|c|}{T}      
    & \multicolumn{1}{c|}{+ 3.54 \%}  
    & \multicolumn{1}{c|}{$-$14.74 + 6.18}
    & \multicolumn{1}{c||}{$-$8.99 + 5.06}
    & \multicolumn{1}{c|}{$-$3.94 \%} 
    & \multicolumn{1}{c|}{+\textbf{3.49} + \textbf{1.68}}
    & \multicolumn{1}{c||}{$-$2.59 + 6.02}
    & \multicolumn{1}{c|}{+ \textbf{5.06} \%} 
    & \multicolumn{1}{c|}{$-$5.87 + 3.16}
    & \multicolumn{1}{c|}{$-$\textbf{4.04} + \textbf{0.04}}
    \\ \hline
\end{tabular}
}
\vspace{0.5em}
\caption{Performance gain comparisons of design decisions, expressed as median difference values in F1 score (\%), onset and offset error (ms). The top table expresses global gains, whereas the bottom table shows gains whenever spatial dropout (SDo) was not applied. A positive F1 score indicates that the design decision increases performance, whereas negative onset/offset mean or STD errors indicate a more precise fiducial location. Bold values represent best performing approaches.}
\label{tab:resultsgains}
\end{table*}

Following the experiments described in Section \ref{sec:experiments}, the contributions of the specific design decisions were independently tested. The pipeline mainly benefited from the application of SDo regularization approach, reaching improvements of 1.98\%, 3.27\% and 7.21\% F1 score in the detection of the P, QRS and T waves, and a reductions of $-2.79 - 0.83$ ms, $-0.73 \pm 1.43$ ms and $8.06 + 3.90$ ms in onset error and of $-2.07 - 1.18$ ms, $+1.88 \pm 2.90$ ms and $-2.81 - 1.32$ ms in offset errors in the P, QRS and T waves, respectively. Such generalized improvement can also be seen in other design decisions like pre-training the network with low-quality labels and the application of DA. Full results can be visualized in Table \ref{tab:resultsgains}.

Other additions showed unclear and inconclusive upon different executions. Some design decisions, such as MsU, HdC and multi-lead inference, improved results only when SDo was not applied; they showed little improvement or even performance degradation otherwise. Table \ref{tab:resultsgains} (bottom) summarizes the performance impact whenever SDo was not applied. ASPP, however, showed consistent performance degradation for our application and network topology. Other additions such as the type of convolutional block, width and depth of the network showed comparable performance throughout all executions. 
\if\boolsupplementary1
A comprehensive list of all variations can be seen in the Supplementary Materials.
\fi

\subsection{Best performing model}

Both single-lead and multi-lead best performing models feature strong regularization techniques in the shape of pre-training with low-quality labelled data and SDo of 25\%, neither of which involve architectural variations (ASPP, HdC or MsU). 

The best performing single-lead model, in accordance to the results expressed above, consists in a model with 5 levels and 3 blocks per level employing the ``vanilla'' convolutional block, with P, QRS and T wave precisions of 90.12\%, 99.14\% and 98.25\% and recalls of 98.73\%, 99.94\% and 99.88\% for detection. The delineation performance shows errors of $1.54 \pm 22.89$ ms, $-0.07 \pm 8.37$ ms and $21.57 \pm 66.29$ ms in the onset and of $0.32 \pm 4.01$ ms, $3.64 \pm 12.55$ ms and $4.55 \pm 31.11$ ms in the offset for delineation, with Dice scores of 88.99\%, 92.05\%, 88.40\% for the P, QRS and T waves, respectively. 

The best multi-lead model features 4 levels and 6 blocks per level employing the ``XCeption'' convolutional block, reaching P, QRS and T precisions of 94.17\%, 99.40\% and 96.36\% and recalls of 94.70\%, 99.28\% and 99.09\% for detection. The delineation performance deviated from the ground truth $1.54 \pm 22.89$ ms, $1.54 \pm 22.89$ ms and $1.54 \pm 22.89$ ms in the onset and $4.01 \pm 16.08$ ms, $5.39 \pm 16.77$ ms and $9.93 \pm 46.33$ ms in the offset, reaching Dice scores of 88.19\%, 92.14\%, 89.33\% for the P, QRS and T waves, respectively.

\begin{table*}[!t]
\centering
{
\scriptsize
\begin{tabular}{llcccccccccccccc}
\cline{3-16}
& \multicolumn{1}{l|}{}
    & \multicolumn{7}{c||}{Precision (\%)}
    & \multicolumn{7}{c|}{Recall (\%)}
    \\ \cline{3-16}
    \multirow{2}{*}{}
    & \multicolumn{1}{l|}{}
    & \multicolumn{1}{c|}{\multirow{2}{*}{\begin{tabular}[c]{@{}c@{}}Single-\\lead\end{tabular}}}
    & \multicolumn{1}{c|}{\multirow{2}{*}{\begin{tabular}[c]{@{}c@{}}Multi-\\lead\end{tabular}}}
    & \multicolumn{1}{c|}{\multirow{2}{*}{\cite{Martinez2004}}}
    & \multicolumn{1}{c|}{\multirow{2}{*}{\cite{Camps2019}}}
    & \multicolumn{3}{c||}{\cite{Sodmann2018}}
    & \multicolumn{1}{c|}{\multirow{2}{*}{\begin{tabular}[c]{@{}c@{}}Single-\\lead\end{tabular}}}
    & \multicolumn{1}{c|}{\multirow{2}{*}{\begin{tabular}[c]{@{}c@{}}Multi-\\lead\end{tabular}}}
    & \multicolumn{1}{c|}{\multirow{2}{*}{\cite{Martinez2004}}}
    & \multicolumn{1}{c|}{\multirow{2}{*}{\cite{Camps2019}}}
    & \multicolumn{3}{c|}{\cite{Sodmann2018}}
\\ \cline{7-9}\cline{14-16}
    & \multicolumn{1}{l|}{}
    & \multicolumn{1}{c|}{}
    & \multicolumn{1}{c|}{}
    & \multicolumn{1}{c|}{}
    & \multicolumn{1}{c|}{}
    & \multicolumn{1}{c|}{10ms}
    & \multicolumn{1}{c|}{50ms}
    & \multicolumn{1}{c||}{150ms}
    & \multicolumn{1}{c|}{}
    & \multicolumn{1}{c|}{}
    & \multicolumn{1}{c|}{}
    & \multicolumn{1}{c|}{}
    & \multicolumn{1}{c|}{10ms}
    & \multicolumn{1}{c|}{50ms}
    & \multicolumn{1}{c|}{150ms}
    \\ \hline
\multicolumn{2}{|l|}{P}
    & \multicolumn{1}{c|}{90.12}
    & \multicolumn{1}{c|}{\textbf{94.17}}
    & \multicolumn{1}{c|}{91.03}
    & \multicolumn{1}{c|}{N/A}
    & \multicolumn{1}{c|}{79.6}
    & \multicolumn{1}{c|}{84.6}
    & \multicolumn{1}{c||}{90.0}
    & \multicolumn{1}{c|}{98.73}
    & \multicolumn{1}{c|}{94.70}
    & \multicolumn{1}{c|}{\textbf{98.87}}
    & \multicolumn{1}{c|}{N/A}
    & \multicolumn{1}{c|}{86.8}
    & \multicolumn{1}{c|}{92.2}
    & \multicolumn{1}{c|}{98.1}
    \\ \hline
\multicolumn{2}{|l|}{QRS}
    & \multicolumn{1}{c|}{99.14}
    & \multicolumn{1}{c|}{99.40}
    & \multicolumn{1}{c|}{\textbf{99.86}}
    & \multicolumn{1}{c|}{N/A}
    & \multicolumn{1}{c|}{93.0}
    & \multicolumn{1}{c|}{98.5}
    & \multicolumn{1}{c||}{99.9}
    & \multicolumn{1}{c|}{\textbf{99.94}}
    & \multicolumn{1}{c|}{99.28}
    & \multicolumn{1}{c|}{99.80}
    & \multicolumn{1}{c|}{N/A}
    & \multicolumn{1}{c|}{92.2}
    & \multicolumn{1}{c|}{97.7}
    & \multicolumn{1}{c|}{99.1}
    \\ \hline
\multicolumn{2}{|l|}{T}
    & \multicolumn{1}{c|}{\textbf{98.25}}
    & \multicolumn{1}{c|}{96.36}
    & \multicolumn{1}{c|}{97.79}
    & \multicolumn{1}{c|}{N/A}
    & \multicolumn{1}{c|}{80.2}
    & \multicolumn{1}{c|}{87.4}
    & \multicolumn{1}{c||}{97.7}
    & \multicolumn{1}{c|}{\textbf{99.88}}
    & \multicolumn{1}{c|}{99.09}
    & \multicolumn{1}{c|}{99.77}
    & \multicolumn{1}{c|}{N/A}
    & \multicolumn{1}{c|}{80.7}
    & \multicolumn{1}{c|}{87.9}
    & \multicolumn{1}{c|}{98.3}
    \\ \hline

&&&&&&&&&&&&&&&\\ \cline{3-16}

& \multicolumn{1}{l|}{}
    & \multicolumn{7}{c||}{Onset Error, ms}
    & \multicolumn{7}{c|}{Offset Error, ms}
    \\ \cline{3-16}
& \multicolumn{1}{l|}{}
    & \multicolumn{1}{c|}{\multirow{2}{*}{\begin{tabular}[c]{@{}c@{}}Single-\\lead\end{tabular}}}
    & \multicolumn{1}{c|}{\multirow{2}{*}{\begin{tabular}[c]{@{}c@{}}Multi-\\lead\end{tabular}}}
    & \multicolumn{1}{c|}{\multirow{2}{*}{\cite{Martinez2004}}}
    & \multicolumn{1}{c|}{\multirow{2}{*}{\cite{Camps2019}}}
    & \multicolumn{3}{c||}{\cite{Sodmann2018}}
    & \multicolumn{1}{c|}{\multirow{2}{*}{\begin{tabular}[c]{@{}c@{}}Single-\\lead\end{tabular}}}
    & \multicolumn{1}{c|}{\multirow{2}{*}{\begin{tabular}[c]{@{}c@{}}Multi-\\lead\end{tabular}}}
    & \multicolumn{1}{c|}{\multirow{2}{*}{\cite{Martinez2004}}}
    & \multicolumn{1}{c|}{\multirow{2}{*}{\cite{Camps2019}}}
    & \multicolumn{3}{c|}{\cite{Sodmann2018}}
    \\ \cline{7-9}\cline{14-16}
    & \multicolumn{1}{l|}{}
    & \multicolumn{1}{c|}{}
    & \multicolumn{1}{c|}{}
    & \multicolumn{1}{c|}{}
    & \multicolumn{1}{c|}{}
    & \multicolumn{1}{c|}{10ms}
    & \multicolumn{1}{c|}{50ms}
    & \multicolumn{1}{c||}{150ms}
    & \multicolumn{1}{c|}{}
    & \multicolumn{1}{c|}{}
    & \multicolumn{1}{c|}{}
    & \multicolumn{1}{c|}{}
    & \multicolumn{1}{c|}{10ms}
    & \multicolumn{1}{c|}{50ms}
    & \multicolumn{1}{c|}{150ms}
    \\ \hline
\multicolumn{1}{|l|}{\multirow{2}{*}{P}}
    & \multicolumn{1}{l|}{Mean}
    & \multicolumn{1}{c|}{\textbf{1.54}}
    & \multicolumn{1}{c|}{-1.72}
    & \multicolumn{1}{c|}{2.0}
    & \multicolumn{1}{c|}{\multirow{2}{*}{N/A}}
    & \multicolumn{1}{c|}{\multirow{2}{*}{N/A}}
    & \multicolumn{1}{c|}{\multirow{2}{*}{N/A}}
    & \multicolumn{1}{c||}{\multirow{2}{*}{N/A}}
    & \multicolumn{1}{c|}{\textbf{0.32}}
    & \multicolumn{1}{c|}{4.01}
    & \multicolumn{1}{c|}{1.9}
    & \multicolumn{1}{c|}{\multirow{2}{*}{N/A}}
    & \multicolumn{1}{c|}{\multirow{2}{*}{N/A}}
    & \multicolumn{1}{c|}{\multirow{2}{*}{N/A}}
    & \multicolumn{1}{c|}{\multirow{2}{*}{N/A}}
    \\ \cline{2-5}\cline{10-12}
\multicolumn{1}{|l|}{}
    & \multicolumn{1}{l|}{STD}
    & \multicolumn{1}{c|}{22.89}
    & \multicolumn{1}{c|}{17.83}
    & \multicolumn{1}{c|}{\textbf{14.8}}
    & \multicolumn{1}{c|}{}
    & \multicolumn{1}{c|}{}
    & \multicolumn{1}{c|}{}
    & \multicolumn{1}{c||}{}
    & \multicolumn{1}{c|}{15.99}
    & \multicolumn{1}{c|}{16.08}
    & \multicolumn{1}{c|}{\textbf{12.8}}
    & \multicolumn{1}{c|}{}
    & \multicolumn{1}{c|}{}
    & \multicolumn{1}{c|}{}
    & \multicolumn{1}{c|}{}
    \\ \hline
\multicolumn{1}{|l|}{\multirow{2}{*}{QRS}}
    & \multicolumn{1}{l|}{Mean}
    & \multicolumn{1}{c|}{\textbf{-0.07}}
    & \multicolumn{1}{c|}{-3.83}
    & \multicolumn{1}{c|}{4.6}
    & \multicolumn{1}{c|}{-2.6}
    & \multicolumn{1}{c|}{\multirow{2}{*}{N/A}}
    & \multicolumn{1}{c|}{\multirow{2}{*}{N/A}}
    & \multicolumn{1}{c||}{\multirow{2}{*}{N/A}}
    & \multicolumn{1}{c|}{3.64}
    & \multicolumn{1}{c|}{5.39}
    & \multicolumn{1}{c|}{\textbf{0.8}}
    & \multicolumn{1}{c|}{4.4}
    & \multicolumn{1}{c|}{\multirow{2}{*}{N/A}}
    & \multicolumn{1}{c|}{\multirow{2}{*}{N/A}}
    & \multicolumn{1}{c|}{\multirow{2}{*}{N/A}}
    \\ \cline{2-6}\cline{10-13}
\multicolumn{1}{|l|}{}
    & \multicolumn{1}{l|}{STD}
    & \multicolumn{1}{c|}{8.37}
    & \multicolumn{1}{c|}{14.64}
    & \multicolumn{1}{c|}{\textbf{7.7}}
    & \multicolumn{1}{c|}{10.8}
    & \multicolumn{1}{c|}{}
    & \multicolumn{1}{c|}{}
    & \multicolumn{1}{c||}{}
    & \multicolumn{1}{c|}{12.55}
    & \multicolumn{1}{c|}{16.77}
    & \multicolumn{1}{c|}{\textbf{8.7}}
    & \multicolumn{1}{c|}{15.2}
    & \multicolumn{1}{c|}{}
    & \multicolumn{1}{c|}{}
    & \multicolumn{1}{c|}{}
    \\ \hline
\multicolumn{1}{|l|}{\multirow{2}{*}{T}}
    & \multicolumn{1}{l|}{Mean}
    & \multicolumn{1}{c|}{21.57}
    & \multicolumn{1}{c|}{\textbf{19.10}}
    & \multicolumn{1}{c|}{\multirow{2}{*}{N/A}}
    & \multicolumn{1}{c|}{\multirow{2}{*}{N/A}}
    & \multicolumn{1}{c|}{\multirow{2}{*}{N/A}}
    & \multicolumn{1}{c|}{\multirow{2}{*}{N/A}}
    & \multicolumn{1}{c||}{\multirow{2}{*}{N/A}}
    & \multicolumn{1}{c|}{4.55}
    & \multicolumn{1}{c|}{9.93}
    & \multicolumn{1}{c|}{\textbf{-1.6}}
    & \multicolumn{1}{c|}{\multirow{2}{*}{N/A}}
    & \multicolumn{1}{c|}{\multirow{2}{*}{N/A}}
    & \multicolumn{1}{c|}{\multirow{2}{*}{N/A}}
    & \multicolumn{1}{c|}{\multirow{2}{*}{N/A}}
    \\ \cline{2-4}\cline{10-12}
\multicolumn{1}{|l|}{}
    & \multicolumn{1}{l|}{STD}
    & \multicolumn{1}{c|}{\textbf{66.29}}
    & \multicolumn{1}{c|}{66.51}
    & \multicolumn{1}{c|}{}
    & \multicolumn{1}{c|}{}
    & \multicolumn{1}{c|}{}
    & \multicolumn{1}{c|}{}
    & \multicolumn{1}{c||}{}
    & \multicolumn{1}{c|}{31.11}
    & \multicolumn{1}{c|}{46.33}
    & \multicolumn{1}{c|}{\textbf{18.1}}
    & \multicolumn{1}{c|}{}
    & \multicolumn{1}{c|}{}
    & \multicolumn{1}{c|}{}
    & \multicolumn{1}{c|}{}
    \\ \hline
\end{tabular}
}
\vspace{0.5em}
\caption{Precision and recall ($\%$; top) and onset and offset errors (mean, M $\pm$ standard deviation, SD; bottom) of our best performing single-lead model as compared to other approaches. Sodmann \textit{et al}. \cite{Sodmann2018} report results for different window sizes (10, 50, 150 ms), considering a true positive if their prediction is contained within the window. N/R stands for ``not reported''.}
\label{tab:resultsbestmodels}
\end{table*}

\begin{table*}[!b]
    \scriptsize
    \centering
    \begin{tabular}{l|c|c|c|c|c|c|c|}
    \cline{2-8}
    & \multicolumn{7}{c|}{QRS width estimation error (ms)} \\ \cline{2-8} 
    & \begin{tabular}[c]{@{}c@{}}MIT-BIH\\ Arrhythmia\end{tabular} & \begin{tabular}[c]{@{}c@{}}MIT-BIH\\ ST Change\end{tabular} & \begin{tabular}[c]{@{}c@{}}MIT-BIH\\ Supraventricular\\ Arrhythmia\end{tabular} & \begin{tabular}[c]{@{}c@{}}MIT-BIH\\ Sinus Rhythm\end{tabular} & European ST-T & Sudden Death & \begin{tabular}[c]{@{}c@{}}MIT-BIH\\ Long-Term ECG\end{tabular} \\ \hline
    \multicolumn{1}{|l|}{Best single-lead} & 14.73 & 6.72 & 7.52 & 6.57 & 8.35 & 14.64 & 10.12 \\ \hline
    \multicolumn{1}{|l|}{Best multi-lead}  & 19.53 & 9.46 & 13.5 & 10.41 & 14.45 & 22.54 & 14.96 \\ \hline
    \end{tabular}
    \vspace{0.5em}
    \caption{QRS width estimation error (ms) for different conditions represented in the database.}
    \label{tab:resultsQRSwidth}
\end{table*}

\begin{figure}[!ht]
    \centering
    \includegraphics[width=0.985\linewidth]{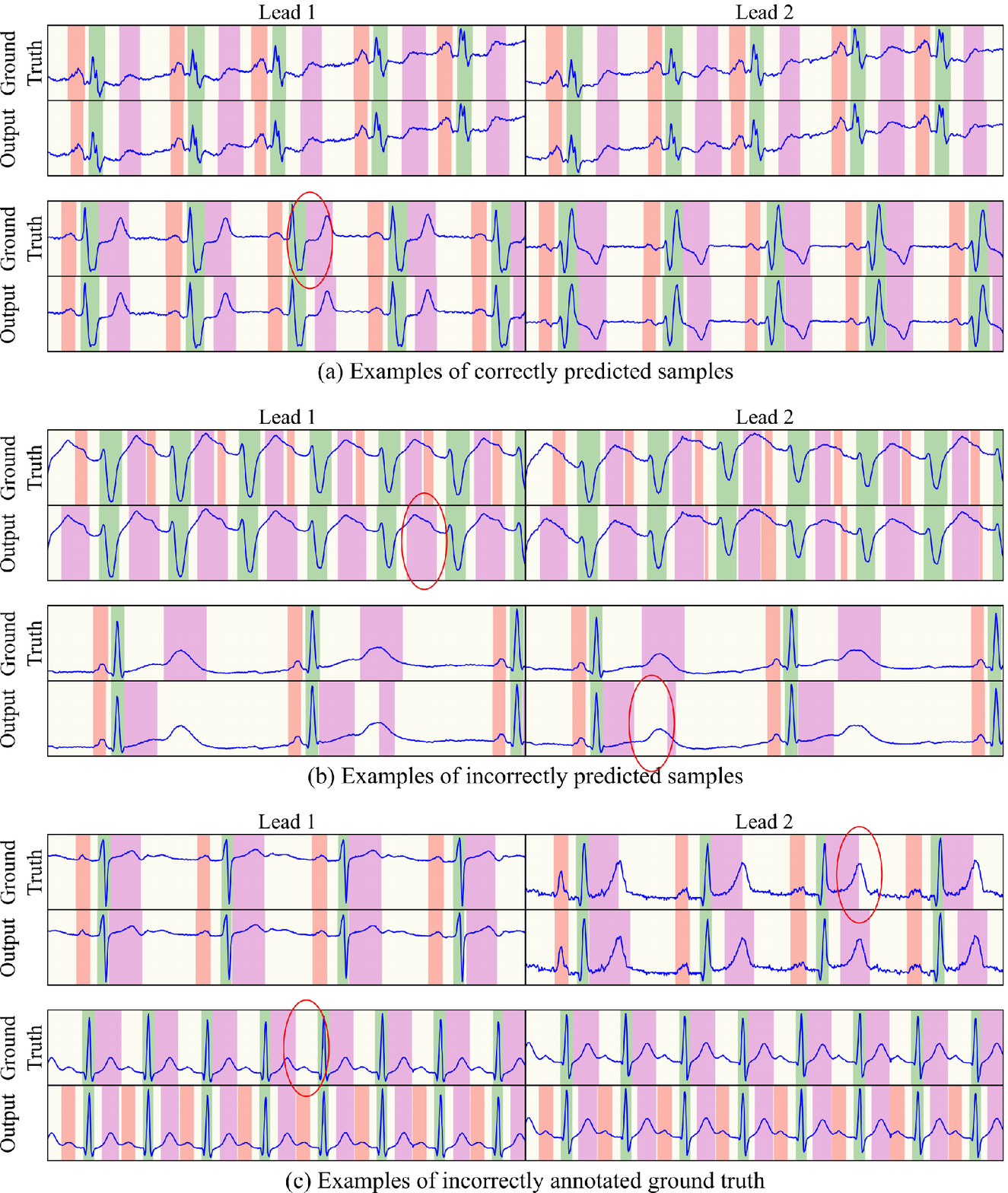}
    \caption{\label{fig:examples} 
    Examples of our proposed single-lead approach, showing the ground truth for that signal (top) and the predicted mask (bottom) for leads 1 (left) and 2 (right). (a) Examples of correctly predicted samples, depicting samples from sudden cardiac death (top) and ST change (bottom). (b) Examples of incorrectly predicted samples, featuring fused T and P waves (top) and severe bradycardia (bottom). (c) Examples of incorrectly annotated ground truth, demonstrating incorrect T offset location (top) and missed P waves (bottom). Red mask: P wave. Green mask: QRS wave. Magenta mask: T wave. Representative examples have been encircled (red) for easier visualization.}
\end{figure}

The optimal network configuration for both single- and multi-lead is detailed in Table \ref{tab:resultsbestmodels}, whereas Figure \ref{fig:examples} depicts several samples from the single-lead and multi-lead approaches. Table  \ref{tab:resultsQRSwidth} details the performance of the model for QRS width estimation for the best single- and multi-lead strategies since it is one of the most used ECG-based indices for clinical decisions. The obtained QRS width estimation error ranges from 6.57 ms (sinus rhythm) to 14.73 ms (arrhythmia) for single-lead, and from 9.46 ms (ST change) to 22.54 ms (sudden death) for multi-lead scenarios.

\section{Discussion}\label{sec:discussion}

\subsection{General observations}

Deep learning techniques usually show notable improved performance upon classical approaches for supervised tasks given sufficient training data \cite{SanchezMartinez2019} and can be successfully used to improve and automate laborious tasks in the medical domain such as image (or signal) segmentation, leading to more efficient workload distribution by augmenting decision-making information \cite{Chen2017}. Under this context, this work presents a FCN-based approach for ECG delineation by framing the problem as a segmentation task. Our work features good detection performance, in both single-lead and multi-lead scenarios, even correcting in some cases systematic errors in the manual delineation (Figure \ref{fig:examples}-c). Our best performing approach, single-lead, performed on par with state-of-the-art delineation approaches (Table \ref{tab:resultsbestmodels}-top, \cite{Martinez2004}) with differences in precision and recall of less than 1\%, even considering that our network was trained using small amounts of annotated data and that the impact of a possible over-adjustment of the rule-based algorithm to the development dataset on their generalizability remains untested \cite{Minchole2019}.

\begin{figure}[!t]
    \centering
    \includegraphics[width=1\linewidth]{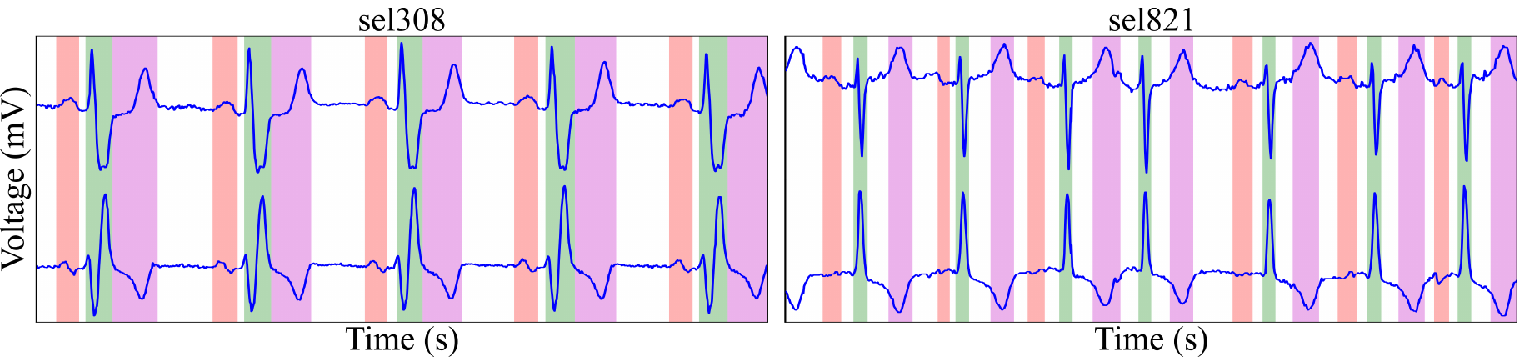}
    \caption{\label{fig:inconsistentTon} 
    High-quality ground truth depicting T onset inconsistencies in two different recordings. P wave (red), QRS wave (green), T wave (magenta).}
\end{figure}

The network, however, produces higher delineation errors when compared to the state-of-the-art when localizing the exact fiducial locations, especially in the T wave (Table \ref{tab:resultsbestmodels}-bottom). The multi-lead approach showed consistently worse performance at delineation, with up to 15 ms difference in the T wave offset. These errors are, however, consistent with the ML-based literature employing the QT database \cite{Hedayat2018, Sodmann2018}, indicating an intrinsic bias in data itself, also noted by other authors \cite{Martinez2010}. We hypothesize that these large differences are due the small amount of annotated data (105 recordings), to the high variability in the represented pathologies (6 pathologies alongside sinus rhythm) and to the inconsistencies of the annotated data. The T wave onset is worthy of special mention as inconsistent criteria were used throughout the database for its markup, as shown in Figures \ref{fig:inconsistentTon} and \ref{fig:examples}-c, with some authors explicitly excluding its performance metrics when reporting their results \cite{Martinez2004}. The network also produces systematic errors in some waves by predicting rise and falls as independent waves (Figure \ref{fig:examples}-b), which we associate to the lack of a wider variety of cardiac conditions and to inconsistencies in the high-quality ground truth. Alternatively, explicit connectivity priors could be included in the data representation \cite{Graves2006}.

Lastly, the inspected model variations showed inconsistent performance gains. Only explicit regularization strategies such as SDo, pre-training on low-quality labels and DA consistently improved overall detection and delineation performance, increasing F1 scores and decreasing onset and offset errors. Architectural variations, namely MsU and HdC, improved the network's performance only if SDo was not applied, whereas ASPP did not demonstrate a clear and consistent positive effect on model performance. We hypothesize that MsU and HdC impose constraints on the channel's structure, which might be mitigated whenever SDo, an explicit regularizer also operating on the channels, is applied. ASPP, on its behalf, increments the model capacity relating distant elements, which indicates that a base U-Net might suffice for this task.

\subsection{Comparison with state-of-the-art approaches}

Our work shows state-of-the-art detection and delineation performance, with values on par with DSP-based methods while presenting competitive advantages and increased performance over DL-based works. DSP-based approaches, like Martínez \textit{et al}. \cite{Martinez2004} provide a high delineation performance and are considered state-of-the-art. However, these algorithms require laborious rule re-calibration when extended to other morphologies. Our approach provides similar results to DSP-based methods while applying cross-validation with strict subject-wise splitting for ensuring generalization.

DL-based approaches, on their behalf, use a variety of methodologies, providing a good framework for comparison; especially given they are dealing with ``small data''. However, it is noteworthy that the compared literature does not commonly details how train/test splitting is made, leading to potentially misleading model performance, as noted by \cite{Faust2018}.

Although limited bibliography of CNN-based methods exists, these compare unfavourably to our FCN-based approach. Camps \textit{et al}. \cite{Camps2019} only produced delineation of the QRS, neglecting P and T waves, and attaining delineation performance of $-2.6 \pm 10.8$ ms and $4.4 \pm 15.2$ ms for the QRS onset and offset. The authors do not report precision or recall metrics, difficulting direct comparison. Sodmann \textit{et al}. \cite{Sodmann2018} attempted at directly predicting the fiducial's sample of occurrence ($w$) through fully convolutional layers. However, these layers are not translation invariant. The authors also excluded 23 recordings of the QT database. In addition, their work suffers from performance pitfalls, achieving differences in performance up to 10\% difference with respect to DSP-based approaches even with large ($\sim$50 ms) tolerance windows, while disregarding detections with large error ($>$250 ms). A summary of their results can be seen in Table \ref{tab:resultsbestmodels}.

A single recurrent formulation employing LSTM has been proposed in the literature by Abrishami \textit{et al}. \cite{Hedayat2018}. Their work did not, however, provide measurements of the delineation performance. Additionally, recurrent neural networks have optimization problems \cite{Pascanu2013,Bai2018}, which might explain their relatively low precision for the QRS and T waves (94\% and 90\%, respectively) and overall low recall (91\%, 94\% and 91\% for the P, QRS and T waves, respectively). On the other hand, the authors did perform subject-wise splitting.

Lastly, Tison \textit{et al}. \cite{Tison2019} recently published a U-Net based ECG delineation methodology for the delineation of 12-lead ECGs, similar to our initial attempt published in \cite{Jimenez-Perez2019}, which was the basis of this work. Tison \textit{et al}. presented an asymmetric U-Net featuring an appended structure at the base level for producing late 12-lead fusion and direct 8-fold upsampling from level 5 to level 2. The authors reported a high Dice score (P wave: 91 $\pm$ 3 \%; QRS wave: 94 $\pm$ 4 \%; T wave: 92 $\pm$ 5 \%) for a private ECG database. However, the authors benefited from enhanced context of the 12 leads in each recording, and discarded recordings that had large error in a downstream task. The authors also cite the crucial role of a HMM-based post-processing step for refining the results. These details could point to overfitting, given that the authors do not apply strong regularization strategies. Their choice of network topology might also hinder the back-propagation algorithm due to the lack of convolutional blocks and skip connections at levels 3 and 4. Finally, their work is also restricted to analysis on sinus rhythm beats, compromising its generalizability to pathological beats, which are harder to correctly delineate (Table \ref{tab:resultsQRSwidth}). Although direct comparison is difficult in this case, our network does not require post-processing, has been tested against a standard database and was based on a well-founded architecture, allowing for a more direct comparison with other DL works in the biomedical engineering domain.

\subsection{Learning points}

Besides the competitive delineation performance, we learnt several lessons for processing ECG data with DL-based techniques. Firstly, when working with ECG data, strong regularization techniques such as SDo and DA are of utmost importance, as the network easily overfitted the training data on the first epoch while stagnating the validation loss. Secondly, pre-training the network on low-quality tags provides increased performance, plausibly due to the increased input data variability. Given the scarcity of manually annotated data in the QT database, low-quality data can give sufficient samples to learn better abstractions, acting as a regularizer via data \cite{Kukavcka2017}. Finally, the application of ECG-based DA methodologies seemed to increase overall performance of the network, when access to a larger database or to low-quality labelling is not possible.

An especially interesting result is drawn from the comparison of the model performance when producing single-lead and multi-lead predictions. The multi-lead fiducial computation suffers $1.89\%$ and $4.03\%$ drops in T wave precision and P wave recall, as well as large (up to $15.22$ ms of difference) in onset and offset standard deviations. These gaps are partially due to the employed evaluation method, which compares the ground truth to the best predicted fiducial, irrespective of the lead on which it has been produced. This methodology, adapted from \cite{Martinez2004} for comparison with DSP-based methods, is a double-edged sword: while it decouples the performance of the delineation to the specific lead fusion strategy, it also masks the error it would inevitably produce. A second reason for the difference in performance is that the multi-lead scenario has half the samples to learn a representation from an input space that is doubly as large, as two leads are used as input.

It is noteworthy that the model's performance degraded consistently at higher capacity models (6 and 7 levels of depth and 4 blocks per level) whenever SDo was not applied, but performed very similarly to other model definitions when it was. An explanation for this is the great imbalance between the network's capacity and the small amount of annotated data, making it more plausible to fall into local minima before regularization. In this sense, one of the most interesting pointers is that best performing model inclusions are those which impose prior knowledge of the data, such as SDo, DA or pre-training, pointing that regularization decisions can be more effective than architectural modifications.

\subsection{Limitations}

Although our network compares positively to the methods in the literature, some aspects leave room for improvement. The main limitation of this work is the lack of more up-to-date databases for delineation, containing a higher variability for a wider array of pathologies. DL is a data-hungry technique in which, as any other, the extrapolation to unseen scenarios has its limits \cite{Marcus2018}. 
However, despite ECG usually being the first information registered of the patient's cardiac condition, not many large annotated databases for ECG analysis exist. The development of reliable DL-based methods for delineation is, thus, tied to data collection and annotation, either containing a large number of multi-lead or single-lead annotations of very different morphologies. With the current existing databases, with approximately 105 different represented ECG morphologies, the current DL-based delineation remains a proof of concept. Other approaches to alleviate this data scarcity would be to develop semi-supervised approaches for making use of large, unannotated databases, implicitly enhancing the learnt representation \cite{Cheplygina2019}, although the usefulness of semi-supervised methods remains unclear \cite{Oliver2018}.

Besides the apparent shortcomings of the existing delineation databases, some improvements could be made in the architecture. One possibility would be to explicitly model temporal dependencies in the shape of RNNs \cite{Hochreiter1997} or attention-based models \cite{Bahdanau2014, Vaswani2017}. Another direction for exploration are high-efficiency models, such as MobileNet \cite{Howard2017}, and model compression \cite{Cheng2017}, for their deployment in CPU-only computers. Finally, novel regularization strategies, further imposition of domain-specific priors, pre-training on similar datasets or developing alternative segmentation losses would further improve performance. Other DA schemes such as varying the heart rate, isoelectric line, specific wave shapes (e.g. voltage or width of P, QRS or T waves within a beat) and the specific SNR and $f_s$ values would also help, but the executions were made to keep an assumable computational budget.

\section{Conclusions}\label{sec:conclusions}

Despite its potential, DL for cardiac signal analysis is not well established in the community \cite{Lyon2018,Minchole2019}. Some influencing factors are the lack of large-scale, quality databases (such as UK BioBank in the imaging community), lack of digital support (many hospitals still print ECGs), lack of per-beat annotated data and the high waveform variability due to pathological conditions, uncertainty in lead positioning, body composition, gender and noisy recordings, among others.

This works contributes by bridging the gap between the imaging and the signal communities for cardiovascular diseases by proving that a DL-based algorithm, properly trained and with an adequate objective function, can provide good delineation with good generalization. This work attempts at helping boost research in the signal-based cardiovascular field by facilitating further downstream tasks. 

With respect to its deployment in a clinical scenario, and although the U-Net is relatively efficient in number of computations, prediction efficiency and model compression would need to be pursued. Another direction for expansion would be to obtain better latent representations either by performing transfer learning from a similar problem in the signal domain or by performing semi-supervised learning or unsupervised pre-training.

\section*{Acknowledgments}

This research was supported by the Secretariat for Universities and Research of the Ministry of Business and Knowledge of the Government of Catalonia and European Social Fund (2017 FI\_B 01008). The TITAN Xp used for this work was donated by the NVIDIA Corporation.

\bibliography{refs.bib}{}

\end{document}